\documentclass[sigconf]{acmart}

\usepackage{booktabs} 
\usepackage{todonotes}
\usepackage{cleveref}
\usepackage{textgreek}

\copyrightyear{2018}
\acmYear{2018}
\setcopyright{acmlicensed}
\acmConference[ESEM '18]{ACM / IEEE International Symposium on Empirical Software Engineering and Measurement (ESEM)}{October 11--12, 2018}{Oulu, Finland}
\acmBooktitle{ACM / IEEE International Symposium on Empirical Software Engineering and Measurement (ESEM) (ESEM '18), October 11--12, 2018, Oulu, Finland}
\acmPrice{15.00}
\acmDOI{10.1145/3239235.3267435}
\acmISBN{978-1-4503-5823-1/18/10}

\acmArticle{4}

\begin{document}
\title{Measuring LDA Topic Stability from Clusters of Replicated Runs}

\author{Mika V. Mantyla}
\orcid{1234-5678-9012}
\affiliation{%
  \institution{M3S, University of Oulu, Finland}
}
\email{mika.mantyla@oulu.fi}

\author{Maelick Claes}
\affiliation{%
   \institution{M3S, University of Oulu, Finland}
}
\email{maelick.claes@oulu.fi}

\author{Umar Farooq}
\affiliation{%
  \institution{M3S, University of Oulu, Finland}
}
\email{umar.farooq@oulu.fi}

\renewcommand{\shortauthors}{M. Mantyla, M. Claes, U. Farooq}

\begin{abstract}
\textbf{Background}: Unstructured and textual data is increasing rapidly and Latent Dirichlet Allocation (LDA) topic modeling is a popular data analysis methods for it. Past work suggests that instability of LDA topics may lead to systematic errors. 
\textbf{Aim:} We propose a method that relies on replicated LDA runs, clustering, and providing a stability metric for the topics.  
\textbf{Method:} 
We generate k LDA topics and replicate this process n times resulting in n*k topics. Then we use K-medioids to cluster the n*k topics to k clusters. The k clusters now represent the original LDA topics and we present them like normal LDA topics showing the ten most probable words. For the clusters, we try multiple stability metrics, out of which we recommend Rank-Biased Overlap, showing the stability of the topics inside the clusters.
\textbf{Results:} We provide an initial validation where our method is used for 270,000 Mozilla Firefox commit messages with k=20 and n=20. We show how our topic stability metrics are related to the contents of the topics.   
\textbf{Conclusions:} 
Advances in text mining enable us to analyze large masses of text in software engineering but non-deterministic algorithms, such as LDA, may lead to unreplicable conclusions. Our approach makes LDA stability transparent and is also complementary rather than alternative to many prior works that focus on LDA parameter tuning. 
\end{abstract}

%
%



 \settopmatter{printfolios=true}
\maketitle
\section{Introduction}\label{sec:intro}
Latent Dirichlet Allocation (LDA) is a topic modeling technique for textual data \cite{blei2003latent} that is widely applied in software engineering   \cite{layman2016topic,agrawal2016wrong,panichella2013effectively,asuncion2010software,thomas2010validating,hemmati2017prioritizing,garousi2016citations,campbell2016latent, sun2016exploring, hindle2012relating, garcia2011enhancing, binkley2014understanding, bird2015art} for different tasks such as requirements engineering \cite{hindle2012relating}, software architecture \cite{garcia2011enhancing}, source code analysis \cite{dit2013feature}, defect reports \cite{layman2016topic}, testing \cite{hemmati2017prioritizing} and to bibliometric analysis of software engineering literature \cite{garousi2016citations,raulamo2015citation}. A survey on topic modelling in software engineering has been conducted \cite{sun2016exploring} and a book  
titled "The art and science of analyzing software data" \cite{bird2015art} devoted a chapter for LDA analysis \cite{campbell2016latent}. Many sources give methodological guidance on how to apply LDA topic modeling in software engineering \cite{agrawal2016wrong, binkley2014understanding,panichella2013effectively}. Given all this, we think it is fair to say that LDA topic modelling is a relevant data analysis technique in empirical software engineering research. 

The quality of the resulting topic model can be evaluated with multiple metrics some inspired by mathematics such as the posterior probability of the topic model given the data \cite{griffiths2004finding}, perplexity of measure in the test data \cite{griffiths2004finding}, or Silhouette coefficient of resulting topics \cite{panichella2013effectively}. Other target metrics are based on empirical observations such as coherence, which measures topic model quality using word co-occurrences in publicly available texts \cite{roder2015exploring}, or stability which investigates similarity of topics between different runs \cite{agrawal2016wrong}. 

Recently, Agrawal et al. \cite{agrawal2016wrong} published a paper titled "What is wrong with topic modeling? And how to fix it using search-based software engineering", where they claimed that the instability of topics is one major shortcoming of this technique. Indeed, studies could result in wrong conclusions if the results are based on instable topics. They proposed using a differential evolution search algorithm to find the input parameters which maximize the topic model stability measured as the similarity of topics between multiple runs. This method reduces instability by finding optimal input parameter settings, but only uses the result of one LDA run which can still have some instable topics.

In this paper, we address the stability of topic models, but rather than optimizing input parameters we propose making stability (or instability) transparent to the user. We achieve this by performing replicated runs of LDA topic modeling and clustering the results. Subsequently, we present the clustering results as any topic modeling results by adding an additional metric of stability. 
Our method is not an alternative to the ones presented by Agrawal et al. \cite{agrawal2016wrong} but additive. Thus, we may use both methods at the same time. However, a benefit of our method is that the topic models may also be optimized towards other targets than stability. For example, a user may choose to optimize the topic model input parameters for coherence \cite{roder2015exploring} or perplexity and still use our approach in the end to provide information about topic stability. 

This paper is structured as follows. In \Cref{sec:method} first we present LDA in more detail and then our method for making the stability transparent. In \Cref{sec:results} we demonstrate our results while \Cref{sec:conc} provides conclusions and discusses future improvements.

\section{Method} \label{sec:method}
\subsection{LDA Topic Modelling}
LDA (Latent Dirichlet Allocation) is a soft clustering algorithm that is ideal for text~\cite{blei2003latent} but also for other purposes such as genetics~\cite{pritchard2000inference} where a relationship between a gene and a genotype can be considered similar to a relationship between a word and a document. 
Given a set of documents, LDA models from what topics this set of documents may have been created from. 
As opposed to hard clustering where each document would be assigned to a single topic only, LDA soft clustering assigns each document a list of topics and probabilities for the topics. 
A topic in LDA is a collection of words and their probability estimates for each topic. 
In order to summarize, after running LDA we have the following.
\begin{itemize}
 \item For all documents $m$ there is a vector $\theta$ which is the topic distribution for that document.
 \item For all topics $k$ there is a vector $\phi$ which is the word distribution for that topic.
\end{itemize}
Before topic generation, LDA requires that we set the input parameters such as the number of topics $k$, and hyper priors $\alpha$ and $\beta$. Past work in software engineering has used different techniques to find optimal input parameters such as genetic algorithms \cite{panichella2013effectively} or differential evolution \cite{agrawal2016wrong}. As pointed out in \Cref{sec:intro}, what is optimal can be measured with many metrics such as perplexity \cite{griffiths2004finding}, stability \cite{agrawal2016wrong}, or coherence \cite{roder2015exploring}.

The stability of a topic model can be defined as the model's ability to replicate its solutions \cite{de2008evaluating}. Instability (the lack of stability) is caused by the non-deterministic nature of Monte-Carlo simulation that is part of the LDA algorithm \cite{agrawal2016wrong}. Past work has shown different stability measures and how to optimize the input parameters to provide a stable topic model \cite{agrawal2016wrong, de2008evaluating, greene2014many}. We think  using the results of a single LDA run, whether optimized for stability or not, is dangerous as perfect stability is impossible to reach. The next section shows a method 
that can be used to make more informed decisions.

\subsection{Transparent Stability}
To make LDA topic stability transparent, we suggest performing replicated LDA runs, clustering the topics, and giving a measure of stability. R-code of our approach is available\footnote{https://github.com/M3SOulu/Measuring-LDA-Topic-Stability}. \Cref{sec:clust_lda}, describes the approach,  \Cref{sec:showing_clus} explains how to show the clusters, and finally \Cref{sec:topic_stability} presents different stability measures. 

\subsubsection{Clustering LDA topics} \label{sec:clust_lda}
As previously described, an LDA topic is a list of words with the probabilities of each word appearing in that topic. When we cluster replicated LDA runs we have $n$ replicated runs, and each run contains $k$ number of topics. Therefore the total number of topics is $t = nk$. Our word list 
is represented by $w$ where $\phi$ is the vector of word distribution for each topic. Thus, we have a topic-word matrix $T$ with dimensions $t\times w$ that we want to cluster back to $k$ clusters as $k$ was the number of topics in our LDA setting. We wanted to take an advantage of the word embeddings produced by GloVE~\cite{pennington2014glove} where our entire word list $w$, which has typically thousands of words, is converted to a word vector space with typically 200-400 elements. It has been shown that in this word vector space, semantically similar words appear close to one another~\cite{pennington2014glove} and we have previously use it for searching software engineering specific synonyms \cite{mantyla2017bootstrapping}. Thus, we form a word vector space with $w$ words and $v$ vectors as matrix $V$ with dimensions ($w\times v$). Then we convert our topic-word matrix $T$ ($t\times w$) to topic-vector matrix $W$ ($t\times v$) via matrix multiplication $T$($t\times w$)$V$($w\times v$). An additional benefit is that the $W$ ($t\times v$) matrix is much smaller than the $T$ ($t\times w$) matrix, resulting in faster clustering. Finally, we use K-medioids clustering to cluster our topics $W$ ($t\times v$) to k topics.

\subsubsection{Showing Clustered LDA topics}\label{sec:showing_clus}
We want to deviate as little as possible from standard LDA topic modeling when presenting the results. We form the list of top ten words for each cluster, in LDA each topic is typically represented by the top ten words. To compute the top 10 words, for a cluster that has multiple topics, we sum up word distributions $\phi$ for all topics in a particular cluster and the top ten words are the ones with the highest sums.  

\subsubsection{Topic stability measure}\label{sec:topic_stability}
At this point, our results would appear like any LDA topic model to the user. However, as we want to give the user transparency to topic stability, we need to add a measure describing topic stability. Obviously, user can investigate each cluster in detail but the topic stability measure can help the user to focus on specific clusters. We propose several measures of topic stability, i.e. whether a set of topics are actually about the same content. When two topics contain the same top 10 words in the same order, then we can think that they are exactly about the same content and should result in a maximum score. On the other hand, any deviations from this should result in a lower score. 

First, \emph{Silhouette} is a well-established measure for cluster validation that considers both how similar each object is to its own cluster (cohesion) and how different it is to other clusters (separation). It has been used in LDA optimization before \cite{panichella2013effectively,mehta2014evaluating}. The average silhouette is produced by the K-medioids clustering performed earlier. However, the cluster separation is not interesting for the user as the user mainly cares about whether a particular cluster has similar elements, i.e. high stability. Furthermore, this measure is based on the absolute values of word probabilities rather than the ranks what are presented to the user.

Second, to model whether the same top words are present and that they are in the same order, we can use \emph{Spearman} rank correlation between the top words of any two topics. Any words that are present in the top word list of one topic, but not the other, are assigned the lowest rank in the other topic. A problem occurs if two topics have the same words but in reverse order, the rank correlation between the topics would be -1 while one would still consider these two topics somewhat similar due to the same top words. Another anomaly is that for two topics with no intersecting top ten words, we would get a better Spearman correlation value than -1 (-0.86).

Third, we can measure \textit{Jaccard} similarity between the top words of any two topics. Extended Jaccard measures have been used in LDA stability task optimization~\cite{agrawal2016wrong,greene2014many}. When two topics have all the same top words, the Jaccard similarity would be 1. On the other hand, the worst case (when all the top words are different) would result in a Jaccard similarity of 0. The undesirable property of the Jaccard similarity is that any variations in ordering would not be reflected in the measure. Obviously, a measure of topic stability should take into account both differences in word intersection and rank of two topics. Luckily, a paper published in 2010 has presented such a measure known as rank-biased overlap (\emph{RBO}) \cite{Webber2010similarity} that seems ideal for LDA topic comparisons. We use the extrapolated version of RBO from R-package Bioconductor\footnote{https://rdrr.io/bioc/gespeR/man/rbo.html} that is computed as follows

\( RBO_{\text{EXT}}(T1, T2, p, d) = \frac{X_d}{d} \cdot p^d + \frac{1-p}{p}\sum_{i=1}^{d} \frac{X_i}{i} \cdot p^i \)

$T1$ and $T2$ are two ranked lists and in our case they are two topics represented by their top words, $d$ is the evaluation depth and in our case it is 10 as we wish to compare top ten words, $X_d$ is the intersection of $T1$ and $T2$ at depth $d$. RBO ranges between 0 and 1. RBO is zero when none of the top words are the same and one when all top words are the same and in the same order. The effect of order, i.e. top-weightedness, is controlled by $p$. When $p$ is 1, the order has no effect and only the intersection is considered. Smaller $p$ gives more weight to order of words. We set $p$ to 0.9 as such value is suggested by RBO authors \cite{Webber2010similarity} and as it seems to offer good balance on impact of different ranks of the top ten words. For an illustration let us consider two topics with the same top ten words but in reverse order. This pair would result in the opposite of Spearman correlation (-1) and Jaccard similarity (1) while the RBO (p=0.9) has value close to the middle of its 0 to 1 range (0.51).

\section{Results}\label{sec:results}
\subsection{Data,  Parameter Tuning, and LDA Runs}
We demonstrate our approach on 271,236 commit messages from  Mozilla Firefox. 
We did some minimal preprocessing by excluding the words appearing fewer than ten times and the ones that appeared in 30\% or more documents. Additionally, we removed common stopwords and individuals' names that appeared as Firefox commit messages are also used for assigning reviewers. We consider this preprocessing very conservative. 

We used a fixed number of topics ($k=20$) and differential evolution algorithm (population=20, $CR = 0.5$, $F = 0.8$, iterations = 10) from the R-package \emph{DEoptim} to optimize for hyper parameters $\alpha$ (0.167) and $\beta$ (0.076) with the target of maximizing perplexity in the holdout set. As previously said, our approach is not affected by hyper parameter tuning so future works may use any training targets or algorithms they see appropriate.

We performed 20 LDA run replications with the R-package \emph{text2vec} (1,000 iterations). This package includes very efficient LDA implementation (WarpLDA~\cite{chen2016warplda}) and training a single model with our data takes less than 2 minutes with a laptop computer. 

\subsection{Stable and Unstable Topics}


\Cref{tab:bestworst} shows the best, worst, and median (eleventh) clusters in terms of LDA topic stability out of the 20 clusters ranked with the RBO stability metric. All four stability metrics are highly correlated with each other in our data and the Pearson correlation range is between 0.91 and 0.98. To demonstrate the details, \Cref{tab:ex_good,tab:ex_mid,tab:ex_bad} shows five topics from the best, worst and median cluster with their five top words. All computations of \Cref{tab:bestworst} were performed with all topics and the top 10 words but \Cref{tab:ex_good,tab:ex_mid,tab:ex_bad} only shows a smaller sample to keep the paper within the four page limit.

\begin{table}
\caption{Best (most stable), median and worst topic clusters} 
\label{tab:bestworst}
\begin{tabular}{rlll}
  \hline
 & Best & Median & Worst \\ 
Silhouette & 0.954 & 0.618 & 0.092 \\ 
  Spearman & 0.953 &  0.255 & -0.295 \\ 
  Jaccard & 0.813 & 0.462 & 0.289 \\ 
  RBO & 0.948 & 0.605 & 0.335 \\ 
  Topics & 20 & 22 & 18 \\ 
  1 & backed & build & update \\ 
  2 & changeset & files & add \\ 
  3 & tree & use & https \\ 
  4 & closed & builds & style \\ 
  5 & backout & file & servo \\ 
  6 & bustage & support & changes \\ 
  7 & failures & add & source \\ 
  8 & changesets & update & patch \\ 
  9 & build & version & css \\ 
  10 & mochitest & windows & support \\ 
   \hline
\end{tabular}
\end{table}

\Cref{tab:bestworst} shows that for the cluster with the best topic stability, all topics have all the words nearly in the same order as the average Spearman rank correlation is very close to 0.95. The average Jaccard similarity in topics is only 0.813, however, we need to remember that if two topics differ by one word of the top ten words this already results in Jaccard similarity of 0.818 (9/11). Manual inspection of the details in \Cref{tab:ex_good} shows that for the five topics shown, the top 5 words appear in the same order. We can further confirm that this is true for all topics in this cluster. Deviations in word rank and occurrence exist in words in places 6 to 10. This topic is about commits that revert (back out) previous changes.

\begin{table}
\caption {Topics of the best topic cluster} 
\label{tab:ex_good}
\begin{tabular}{rlllll}
  \hline
 &Topic 1 &Topic 2 &Topic 3 &Topic 4 &Topic 5 \\ 
  \hline
1 & backed & backed & backed & backed & backed \\ 
  2 & changeset & changeset & changeset & changeset & changeset \\ 
  3 & tree & tree & tree & tree & tree \\ 
  4 & closed & closed & closed & closed & closed \\ 
  5 & backout & backout & backout & backout & backout \\ 
   \hline
\end{tabular}
\end{table}

As expected, the cluster with median stability has a lower topic stability than the best cluster in all stability metrics. We can also notice that number of topics in this cluster is higher than 20, i.e. the number of replicated runs we performed. This means that from a single LDA run, more than one topic is part of this cluster. Our K-medioids clustering took all 400 (20*20) topics as input and we did not try to force it to pick one topic from one LDA run to each cluster. This is something we may want to investigate in the future. \Cref{tab:ex_mid} shows that this cluster is about build file usage or updates. 

\begin{table}
\caption {Topics of the median topic cluster} 
\label{tab:ex_mid}
\begin{tabular}{rlllll}
  \hline
 &Topic 1 &Topic 2 &Topic 3 &Topic 4 &Topic 5 \\ 
  \hline
1 & build & build & build & build & build \\ 
  2 & files & files & files & files & files \\ 
  3 & builds & update & file & file & use \\ 
  4 & file & version & use & use & file \\ 
  5 & use & builds & builds & builds & builds \\ 
   \hline
\end{tabular}
\end{table}

\Cref{tab:ex_bad} shows detailed sample of the worst cluster and we see variations in the word order and occurrence. It appears that this topic is about updates, additions, fixes and changes. Since all these are very common words in a version control context, it is hard to make a meaningful interpretation of the topics in this cluster.


\begin{table}
\caption {Topics of the worst topic cluster} 
\label{tab:ex_bad}
\begin{tabular}{rlllll}
  \hline
 &Topic 1 &Topic 2 &Topic 3 &Topic 4 &Topic 5 \\ 
  \hline
1 & update & update & style & update & style \\ 
  2 & add & https & animation & fix & css \\ 
  3 & version & changes & element & patch & text \\ 
  4 & support & source & patch & x & servo \\ 
  5 & changes & add & https & changes & https \\ 
   \hline
\end{tabular}
\end{table}

\section{Conclusions and Future Work} \label{sec:conc}
Past work in software engineering \cite{agrawal2016wrong} and machine learning \cite{greene2014many} point out that LDA instability may lead to incorrect conclusions and proposes input parameter optimization to alleviate the problem. This paper suggests performing replicated runs, clustering the results and measuring the topic stability. These approach are not alternative but additive. Our approach can be combined with any LDA optimization technique that relies on input parameter optimization. Finally, our approach shows topic stability by providing a metric of topic stability and allowing further investigation of the clusters when desired. 

This paper presents multiple metrics of topic stability in \Cref{tab:bestworst} that are highly correlated with each other in our data set. Based on theoretical metric properties (see \Cref{sec:topic_stability}) we recommend using RBO \cite{Webber2010similarity}. In the future, one should empirically establish what p value setting of RBO metric most accurately matches the user expectation on topic stability as in this paper we only used the default ($p=0.9$). We should also study how the topic clusters can be used in the downstream NLP tasks in software engineering.  Furthermore, to demonstrate our idea we also made other design choices but didn't investigate their impact. For example, we considered only 20 replication runs which might be too little and we only generated 20 LDA topics for each run. We also clustered our topics in the word vector space produced by GLoVe as prior work suggested it would produce better results than clustering in word space. All these choices could be challenged. 

Zeller's 2018 ICSE talk \cite{zeller2018talk} has warned us about the dangers of adding complexity. Our approach adds complexity but eventually hides it behind an RBO metric showing the stability of each topic cluster. If the stability of topics is an issue, users of topics models should be made aware of it but with minimal added complexity as we have tried to do in this paper.
\section*{Acknowledgments}

This work has been supported by Academy of
Finland grant 298020.
\bibliographystyle{ACM-Reference-Format}
\bibliography{99_sample-bibliography}

\end{document}